\documentclass[sigconf]{acmart}
\AtBeginDocument{%
  }

\setcopyright{acmlicensed}
\copyrightyear{2018}
\acmYear{2018}
\acmDOI{XXXXXXX.XXXXXXX}

\acmConference[Conference acronym 'XX]{Make sure to enter the correct
  conference title from your rights confirmation emai}{June 03--05,
  2018}{Woodstock, NY}
\acmISBN{978-1-4503-XXXX-X/18/06}

\usepackage{threeparttable}
\usepackage{multirow}
\usepackage{amsmath}
\usepackage{algorithm}
\usepackage{algorithmic}
\usepackage{balance}
\usepackage{makecell}
\begin{document}

\title{Adapting Amidst Degradation: Cross Domain Li-ion Battery Health Estimation via Physics-Guided Test-Time Training}

\author{Yuyuan Feng}
\email{fengyuyuan01@gmail.com}
\affiliation{%
  \institution{Xiamen University}
  \city{Xiamen}
  \country{China}
}

\author{Guosheng Hu}
\affiliation{%
  \institution{University of Bristol}
  \city{Bristol}
  \country{England}}
\email{huguosheng100@gmail.com}

\author{Xiaodong Li}
\affiliation{%
  \institution{Hong Kong University}
  \city{Hong Kong}
  \country{China}
\email{xiaodlee@foxmail.com}
}

\author{Zhihong Zhang*}
\affiliation{%
 \institution{Xiamen University}
 \city{Xiamen}
 \country{China}
 \email{zhihongzhang@xmu.edu.cn}
 }

\renewcommand{\shortauthors}{Trovato et al.}

\begin{abstract}
Health modeling of lithium-ion batteries (LIBs) is crucial for safe and efficient energy management and carries significant socio-economic implications. Although Machine Learning (ML)-based State of Health (SOH) estimation methods have made significant progress in accuracy, the scarcity of high-quality LIB data remains a major obstacle.
Although existing transfer learning methods for cross-domain LIB SOH estimation have significantly alleviated the labeling burden of target LIB data, they still require sufficient unlabeled target data (UTD) for effective adaptation to the target domain. Collecting this UTD is challenging due to the time-consuming nature of degradation experiments.
To address this issue, we introduce a \emph{practical} Test-Time Training framework, \textbf{BatteryTTT}, which \emph{adapts the model continually using each UTD collected amidst degradation}, thereby significantly reducing data collection time. To fully utilize each UTD, BatteryTTT integrates the inherent physical laws of modern LIBs into self-supervised learning, termed \textbf{Physcics-Guided Test-Time Training}. Additionally, we explore the potential of large language models (LLMs) in battery sequence modeling by evaluating their performance in SOH estimation through model reprogramming and prefix prompt adaptation.  The combination of BatteryTTT and LLM modeling, termed \textbf{GPT4Battery},
achieves state-of-the-art generalization results across current LIB benchmarks. Furthermore, we demonstrate the practical value and scalability of our approach by deploying it in our real-world battery management system (BMS) for 300Ah large-scale energy storage LIBs.

\end{abstract}



\received{20 February 2007}
\received[revised]{12 March 2009}
\received[accepted]{5 June 2009}

\maketitle

\section{Introduction}

\begin{figure}[htbp] 
    \centering 
    \includegraphics[width=\linewidth]{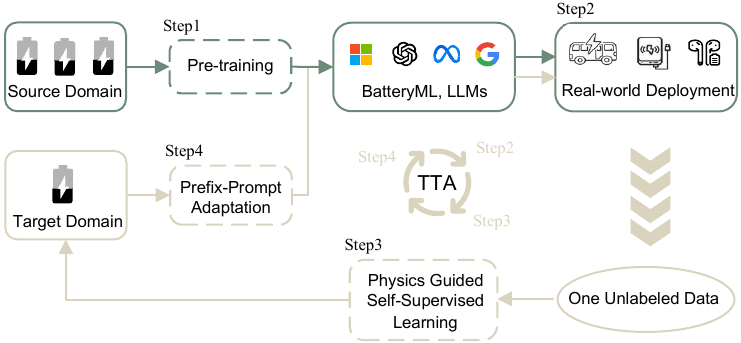} 
    \caption[\protect\footnotemark]{\small Overview of BatteryTTT framework, which consists of three major components: (Step 1) pre-training on experimental datasets; (Step 2) incremental data collection after deployment; and (Steps 3-4) test-time adaptation. Steps 2, 3, and 4 iterate until the LIB retires.}

    \label{overview} 
\end{figure}

The rapid advancements in rechargeable Li-ion batteries (LIBs) have facilitated their widespread use across various sectors, including portable electronics, medical devices, renewable energy systems, and electric vehicles \cite{GENIKOMSAKIS2021119052}. This ubiquity, however, introduces critical challenges associated with capacity degradation and performance evaluation. As an inherently interdisciplinary subject, battery aging modeling has emerged as a fundamental issue at the intersection of battery science and machine learning (ML) \cite{zhangbatteryml, ng2020predicting, severson2019data, roman2021machine}. Accurate State of Health (SOH) estimation for LIBs is crucial not only for ensuring safe and efficient energy management but also for optimizing the design and performance of next-generation batteries, thus having significant socio-economic implications.

With the rapid advancement of ML technology, data-driven SOH estimation models have achieved significant progress in both accuracy and computational efficiency \cite{zhangbatteryml, roman2021machine}.
However, obtaining sufficient training data for LIBs is challenging due to the time-consuming nature of degradation experiments, which typically span months to years. Additionally, precisely labeling this data necessitates additional cycles under controlled temperature and current conditions in the laboratory. Consequently, the scarcity of high-quality LIB data continues to present a major obstacle obstacle in battery aging modeling \cite{zhangbatteryml, severson2019data, lu2023deep}. 

To address the challenge of data scarcity, existing studies have extensively explored transfer learning methods for cross-domain LIB SOH estimation \cite{9410274, TIAN20211521, wang2024lithium, 8870196,lu2023deep}. 
For instance, SSF-WL \cite{wang2024lithium} introduced a self-supervised approach that pre-trains on unlabeled LIB data from a source domain and fine-tunes on a different type of LIB (target domain) using only a small portion of labeled data. This method achieves comparable results using just 30\% of the labeled target data. Another notable study, woDegrad \cite{lu2023deep}, proposed minimizing the domain gap between source and target LIBs by aligning their features in the latent space, eliminating the need for additional labeled target data to estimate SOH.

Unfortunately, while existing studies have significantly alleviated the burden of data labeling, they often overlook the challenge of collecting unlabeled target data (UTD). Due to the time-consuming nature of degradation experiments, gathering sufficient UTD for a typical LIB can take months to years, depending on the battery type. As a result, preparing sufficient UTD for previous algorithms to function effectively also demands a considerable amount of time. For example, migrating a pre-trained woDegrad model \cite{lu2023deep} to the KOKAM dataset \cite{birkl2017oxford}, a widely used LIB dataset, would require a minimum of 8,473 hours to collect enough UTD from KOKAM for effective domain gap alignment, which is hardly practical in real-world applications. To the best of our knowledge, few studies have successfully estimated SOH without relying on a substantial amount of UTD.

\begin{table}[tbp]
  \centering
  \caption{\small Comparison between BatteryTTT and other transfer learning methods regarding the target LIB dataset prerequisites. Note: the time estimates are based on the KOKAM dataset \cite{birkl2017oxford}. }
  \label{assumptions}
  \resizebox{0.95\columnwidth}{!}{
    \begin{threeparttable}
      \begin{tabular}{lccc}
      \toprule
        \textbf{Methods}  & \textbf{woDegrad} & \textbf{SSF-WL} & \textbf{BatteryTTT(Ours)} \\
        \midrule
        \textbf{Unlabeled Data} & 100\% & 30\% & 1 amidst degradation  \\
        \textbf{Labeling}  & 0\% & 30\% &  0\%  \\
        \textbf{Collection Time}  & 8473 hours & 2542 hours & 0 hours \\
        \bottomrule
      \end{tabular}
    \end{threeparttable}
    }
\end{table}

The purpose of this study is to develop a \emph{practical} transfer learning framework for cross-domain LIB SOH estimation that minimizes the reliance on UTDs. Inspired by the Test-Time Training (TTT) \footnote{In the following sections, the terms Test-Time Training (TTT) and Test-Time Adaptation (TTA) may be used interchangeably.} technique from computer vision \cite{sun2020test, gandelsman2022test}, we propose \textbf{BatteryTTT}. Unlike previous methods that require a substantial amount of UTDs collected over time, BatteryTTT adapts the model continually using each \emph{individual UTD collected amidst degradation}. In Table \ref{assumptions}, we compare the prerequisites of the target LIB dataset for BatteryTTT with those of other transfer learning methods. BatteryTTT significantly reduces the amount of required UTD and labeling, offering a more efficient approach compared to existing methods.
To achieve this, BatteryTTT employs a proxy unsupervised task to utilize the UTD for gradient-based model updates. Although some unsupervised methods from existing TTT literature can be applied, such as self-prediction, they are not specifically designed for battery-related tasks, leading to suboptimal performance. To address this, we explore the integration of the inherent physical laws of modern LIBs into self-supervised learning within BatteryTTT, a framework we term \textbf{Physics-Guided Test-Time Training}. This approach leverages the 1-RC Equivalent Circuit Model (ECM) equations to guide the pre-trained model in making accurate self-predictions, leading to improved results. The details of our method and experimental results are discussed in the following sections.

On the other hand, with the rapid advancement of large language models (LLMs) \cite{devlin2018bert, radford2019language, touvron2023llama, radford2018improving} there has been growing interest in exploring their potential for processing \emph{cross-disciplinary sequence data} beyond natural language, including protein sequence prediction \cite{lv2024prollama} and time series analysis \cite{zhou2023one, jin2024position}. 
Pioneering research has validated the efficacy of this paradigm and highlighted the underlying zero-shot generalization capabilities of LLMs across various datasets. However, the application of LLMs to battery sequence modeling remains undiscovered. To address this gap, we evaluate the performance of LLMs for processing cross-domain LIB sequences in this study. Specifically, We employ the concept of model reprogramming to bridge the gap between language and battery modalities. Additionally, we develop a prefix prompt adaptation strategy to efficiently integrate an LLM into our BatteryTTT framework. This combination of strategies, termed \textbf{GPT4Battery}, achieves state-of-the-art generalization results among current LIB benchmarks.

 



By introducing the BatteryTTT framework and GPT4Battery model, we hope this study will inspire the research community to fully leverage advanced AI techniques, such as Test-Time Training (TTT) and large language models (LLMs), for more scenario-fitting AI4Science problems, thereby saving enormous time and accelerating scientific discovery.
For the rest of this paper, 
Sec.\ref{sec:relatedwork} presents background and related works, 
Sec. \ref{Prelimeries} introduces the preliminaries, Sec. \ref{sec:GPT4Battery} describes the methodology of BatteryTTT framework and GPT4Battery model, Sec. \ref{sec:experiments} conducts experiments, and Sec. \ref{sec:conclusion} concludes.

\section{Related Work}\label{sec:relatedwork}

In this section, we introduce the related works in \textit{Data-Driven Battery SOH Estimation} (Sec. \ref{sec:ddbse}),  \textit{Test-Time Adaptation} (Sec. \ref{sec:ttt}) and \textit{LLM for Cross-disciplinary Sequence Modeling} (Sec. \ref{sec:ltsd}), 
 respectively. 
\subsection{Data-Driven Battery SOH Estimation} \label{sec:ddbse}
 Data-driven battery SOH estimation as ascended as a pivotal topic in industrial artificial intelligence and data mining with the widespread adoption of modern LIBs in various applications, bringing a surge in demand for safe and efficient battery management\cite{ng2020predicting, roman2021machine}.
 With the evolution of model architectures within the AI community, algorithms for battery state estimation have also progressed from statistical machine learning methods, such as Random Forest \cite{breiman2001random} and Gaussian Process Regression \cite{rasmussen2010gaussian}, to high-performance deep neural networks, including MLP \cite{haykin1998neural}, LSTM \cite{hochreiter1997long}, and Transformer \cite{zhangbatteryml}.
         
However, to obtain both battery training data and ground-truth labels requires time- and resource-consuming degradation experiments, posing a persistent hurdle in battery aging
modeling \cite{zhangbatteryml, severson2019data, roman2021machine, lu2023deep}.
Noticing this significant issue, researchers in the battery field have tried to use transfer learning methods for generalizable SOH estimation \cite{9410274,TIAN20211521, wang2024lithium, 8870196,lu2023deep}. However, there is a notable gap between current methods' assumptions about having access to the target LIB dataset and the real-world situation, especially regarding the unlabeled target data (UTD). 
while existing studies have significantly alleviated the burden of data labeling, they often overlook the challenge of collecting UTDs. Due to the time-consuming nature of degradation experiments, gathering sufficient UTD for
a typical LIB can take months to years, depending on the battery
type. This cost of time is unacceptable in real-world deployment (as shown in Table \ref{assumptions}). Conversely, BatteryTTT fully utilizes each  individual UTD collected amidst degradation for adaptation to the target domain, thereby significantly reducing data collection time

\subsection{Test-Time Training} \label{sec:ttt}

Test-time Training (TTT)/Adaptation (TTA), also known as one-sample unsupervised domain adaptation, aims to adapt a model trained
on the source domain to the target domain as every unlabeled test sample arrives \cite{sun2020test, liang2024comprehensive}. 
The process of TTA usually involves a self-supervised loss to extract information from the single target domain sample, such as rotation \cite{sun2020test}, mask \cite{gandelsman2022test} and entropy minimization \cite{wang2020tent}. 
Despite the study of various TTA methods, most are designed for (image) classification and cannot be applied to time series regression. 
For instance, Test-time Entropy Minimization (Tent) \cite{wang2020tent} found that the entropy of prediction strongly correlates with accuracy on the
target domain, BACS \cite{zhou2021bayesian}, MEMO \cite{zhang2022memo}, and EATA \cite{niu2022efficient} follow Tent's approach and improve adaptation performance, making them the most representative TTA
methods. Although some unsupervised methods from existing TTT literature can be applied, such as
self-prediction, they are not specifically designed for battery-related
tasks, leading to suboptimal performance.
In this paper, we explore how to incorporate the inherent physics of LIBs into self-supervised learning, resulting in a more natural and powerful approach.


\subsection{LLM for Cross-Disciplinary Sequence Modeling}\label{sec:ltsd}
With the rapid advancement of large language models (LLMs) \cite{devlin2018bert, radford2019language, touvron2023llama, radford2018improving} there has been growing interest in exploring their potential for processing \emph{cross-disciplinary sequence data} beyond natural language, including protein sequence prediction \cite{lv2024prollama} and time series analysis \cite{spathis2023step, zhou2023one, gruver2024large, chang2024llm4ts}. For protein sequence prediction, LLM have showed powerful cross-domain potential against protein language models by the design of vocabulary \cite{ferruz2022protgpt2, doi:10.1073/pnas.2016239118}.
For time series, these efforts have evolved from the initial direct application of large language models (LLMs) to sequence tasks \cite{zhou2023one},  to designing a learned dictionary of prompts to guide inference \cite{cao2023tempo}, to attempting to align the semantic spaces between language and time series modalities \cite{jin2023time, pan2024textbf}.
Although the effectiveness of large language models (LLMs) in handling cross-disciplinary sequence data has been demonstrated, the field of battery research has yet to benefit from this advancement. In this paper, we address this gap by repurposing an LLM for State of Health (SOH) estimation through model reprogramming. Additionally, we evaluate the generalization improvements achieved by adapting an LLM for cross-battery SOH estimation.



\section{Preliminaries} \label{Prelimeries}

\subsection{Battery SOH Definition}
As batteries undergo repeated charge and discharge cycles, their capacity gradually declines due to aging, which leads to performance degradation and potential safety issues. The State of Health (SOH) quantifies the battery's remaining capacity relative to its initial capacity when new. Specifically, if we denote the nominal capacity of the LIB as $C_{norm}$ and the full discharge capacity in the current cycle as $C_{full}$
, the SOH is defined as the ratio of these two values, expressed as a percentage:

\begin{equation}\mathrm{SOH}=\frac{C_{\mathrm{full}}}{C_{\mathrm{nom}}}\times100\%.\end{equation}

LIBs are typically considered to have reached the end of their life cycle when their SOH drops to approximately 75\%.

\subsection{Feature Engineering}
In this study, we utilize \emph{QdLinear} \cite{severson2019data}, a degradation feature derived from the linear interpolation of the voltage-capacity curve during charge cycles, to map the relationship with SOH. This feature is widely recognized and employed by mainstream SOH estimation algorithms \cite{zhangbatteryml, ng2020predicting, severson2019data, lu2023deep}.

\subsection{Problem Definition}

We formalize the problem of cross-domain LIB SOH estimation as follows. Given a well-curated battery dataset from the source domain, denoted as \(\mathcal{S} = \{(x_1, y_1), (x_2, y_2), \ldots, (x_S, y_S)\}\), where \(\mathbf{X} \in \mathbb{R}^{1 \times T}\) represents the extracted \emph{QdLinear} feature with \(T\) time steps, and \(y\) denotes the corresponding SOH label. This source set contains \(S\) labeled lifelong samples. In contrast, for a different battery type in the target domain, we can acquire \emph{only one} unlabeled feature at a time after real-world deployment, represented as \(\mathcal{T} = \{x_1, x_2, \ldots, x_T\}\). Our objective is to estimate each corresponding target label \(y_t\), with \(y_1\) being considered as SOH = 100\% for a new battery.

\section{Methodology}\label{sec:GPT4Battery}

In this section, we present the methodology of our framework. After providing a systematic overview in Section \ref{3.2}, we focus on two key innovations: Physics-Guided Self-Supervised Learning (PG-SSL) and Prefix Prompt Adaptation (PPA), which are detailed in Sections \ref{3.3} and \ref{3.4}. In Section \ref{3.5}, we explain how existing State of Health (SOH) estimation models are integrated into the BatteryTTT framework for cross-domain transfer learning, alongside the exploration of Large Language Models (LLMs) for battery sequence modeling.

\subsection{System Overview} \label{3.2}

Figure \ref{overview} depicts an overview of the BatteryTTT framework, which is composed of three major components: \emph{pre-training on experimental datasets}, \emph{incremental data collection in real-world deployment}, and \emph{test-time adaptation}. 
Firstly, we utilize experimental datasets (source domain) to train a pre-trained model for SOH estimation. We then deploy this pre-trained model to real-world devices, such as the Battery Management System (BMS) of an electric car or a mobile phone. The BMS individually collects unlabeled test data during usage, and upon receiving a sample, we conduct a \emph{test-time training} process to adapt the pre-trained model to this different type of LIB (target domain). Specifically, the test-time adaptation process consists of two steps: PG-SSL to construct an unsupervised loss from the unlabeled sample, and PPA to adapt the pre-trained model to this new domain in a parameter-efficient manner. After learning from incoming data, the adapted model is ready to make a prediction.
This process operates continually until the LIB retires.

\subsection{Physics-Guided Self-supervised Learning} \label{3.3}

\begin{figure}[htbp]
  \centering
    \includegraphics[width=\linewidth]{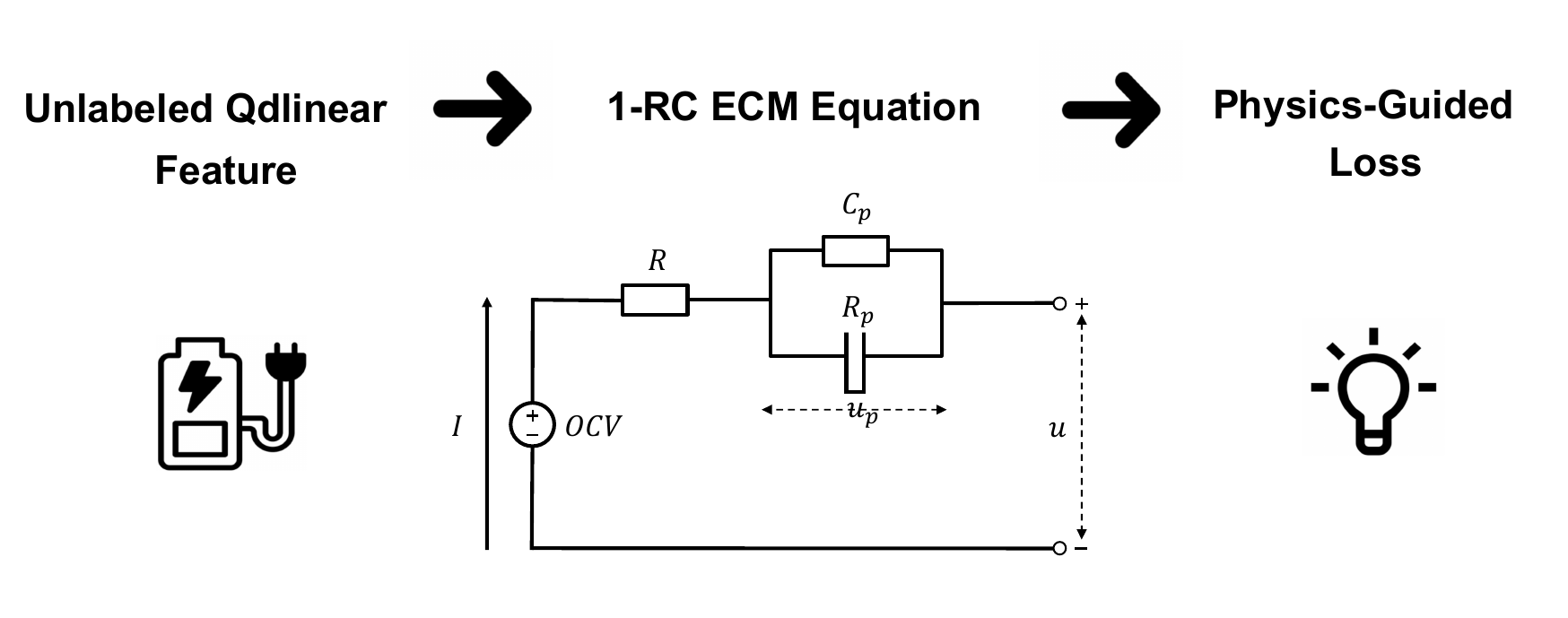}
  \caption{Transform an arbitrary unlabeled Qdlinear feature into a Physics-Guided loss.}
  \label{ECM}
\end{figure}\label{3.2PGSSL}

In this subsection, we elaborate on how to fully exploit the inherent physical laws of a \emph{single} LIB sequence to improve TTA performance through the design of physics-guided self-supervised learning (PG-SSL). Specifically, we first describe how to use the 1-RC ECM equation to transform an arbitrary unlabeled Qdlinear feature into a physics-guided loss (PGLoss). We will then explain how minimizing this PGloss facilitates the estimation of the LIB SOH.

\subsubsection{Thevenin's Equivalent Circuit Model (ECM) of LIB}

The equivalent circuit model (ECM) is widely used battery model to describe the electrical behavior of the battery in terms of voltages, currents, resistances and capacitances \cite{xiong2018towards} \cite{tran2021comprehensive}. The first order Thevenin model is thought to be accurate and adequate to model the condition of the battery, and at the same time simple and computationally efficient \cite{xu2022physics}.
The $OCV$ is represented by an ideal voltage source of the battery. $R$ accounts for the internal ohmic resistance. The parallel $RC$-branch, comprising $R_P$ and $C_P$, is used to model battery polarization effect. $u$ and $I$ denotes the terminal voltage and current that can be collected in use.
Based on Kirchhoff’s law, the electrical behavior of the battery can be characterized as physical equations as:

\begin{equation}
    u_{OCV} = u_R + u_p + u
\end{equation}

\begin{equation}
    \frac{R+R_p}{C_p R_p}I + \frac{1}{C_p R_p}u + \dot{u} = 0
\end{equation}

We define $\theta_1 = \frac{R+R_p}{C_p R_p}$ and $\theta_2 = \frac{1}{C_p R_p}$. Following recent works, the coeffients are functions of temperature. The state function becomes:

\begin{equation}
    \theta_1(T)I + \theta_2(T)u + \dot{u}= 0 \label{ecm}
\end{equation}

This implies that the terminal voltage $u$, current $I$ and temperature should, in principle, follow the ordinary differential equation (ODE) functions representing the battery's physical state, as described in Equation \ref{ecm}. By querying the real-time current and temperature value from the BMS, we can transform an arbitrary unlabeled Qdlinear feature into a PGLoss in an unsupervised manner.

\subsubsection{Generate a complete Qdlinear feature and supervise it with PGLoss} 

\begin{figure}[tbp] 
    \centering 
    \includegraphics[width=\linewidth]{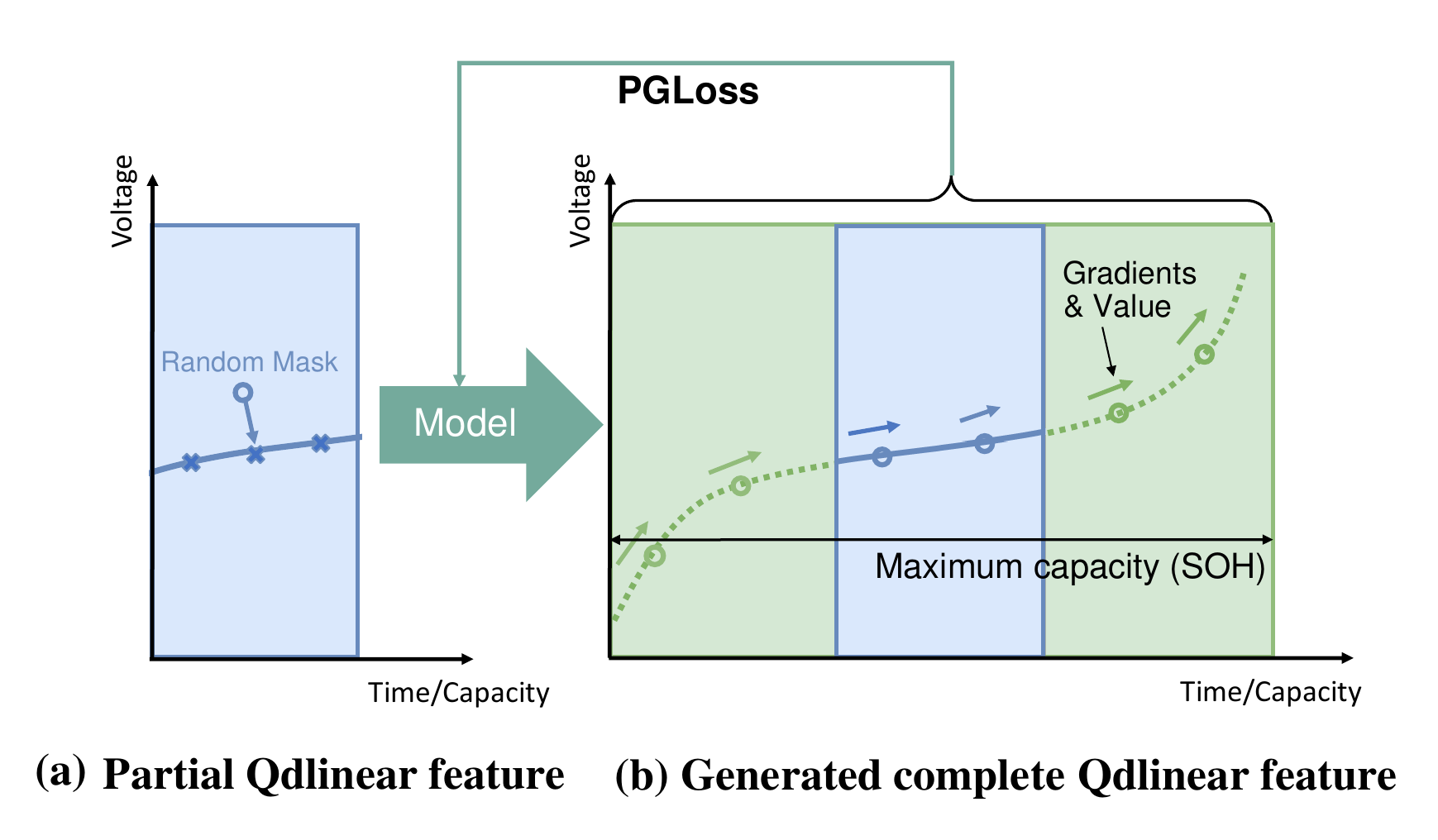} 
    \caption[\protect\footnotemark]{Guide the pre-trained model to generate a complete Qdlinear feature curve and supervise it with Physics-Guided loss.} 
    \label{guiding} 
\end{figure}

In the filed of battery research, a complete Qdlinear curve spanning from the lower to the upper voltage limits can describe LIB's aging mode and therefore can theoretically identify the accurate state of health \cite{yang2021voltage} \cite{chen2022novel} \cite{tian2021deep} according to its definition\footnote{This complete Qdlinear (voltage-capacity) curve requires additional cycles under constrained temperature and current conditions in the laboratory and infeasible at deployment. The unlabeled Qdlinear feature we obtained at use is always a partial of it.}.
Inspired by this, we want to design the objective by guiding the pre-trained model to generate a complete Qdlinear feature from the given partial one, which means to generate Figure \ref{guiding} (b) from (a).

Specifically, given the input (partial) Qdlinear feature curve $\mathbf{x} \in\mathbb{R}^{1\times T}$, we use the pre-trained model to generate a complete one $\mathbf{\hat{x}} \in\mathbb{R}^{1\times T'}$, where $T' > T$. This complete voltage-capacity curve is then supervised with PGLoss (as shown in Figure \ref{guiding}). Additionally, we randomly mask a portion of $\mathbf{x}$ to promote representation learning. In general, our objective can be formulated in the form of:

\begin{equation}
    \mathcal{L}_{\mathrm{PG-SSL}} = \|\hat{\mathbf{x}}[0:T]-\mathbf{x}\|_F^2 + \lambda \|\theta_1 I + \theta_2\hat{u} + \hat{\dot{u}}= 0\|_F^2  \label{pg-ssl}
\end{equation}

Here, $\mathbf{\hat{x}}[0:T]$ represents the overlap between the generated complete voltage-capacity curve and the given $\mathbf{x}$. $\hat{u}$ denotes the generated complete voltage-time curve. The parameters $\theta_1$ and $\theta_2$ are associated with temperature, and $I$ denotes the current; all of these values can be queried through the BMS during deployment. We will empirically demonstrate that the design of Physics-Guided PG-SSL is highly suitable for our SOH estimation task and leads to better performance than simple self-prediction, as shown in the experimental section.

\subsection{Prefix Prompt Adaptation} \label{3.4}

In this subsection, we introduce Prefix Prompt Adaptation (PPA), a method that reduces the dimension of the solution space for easier optimization of test-time adaptation. Specifically, inspired by the demonstrated effectiveness of continuous prompt learning in the field of deep model fine-tuning \cite{jia2022visual, bahng2022exploring}. We add a small number of soft prompts as a prefix to the embedded input context for test-time updating, while keeping all other model parameters frozen. In this way, the dimension of learnable model parameters shall be significantly reduced and thus enabling the practical deployment of TTA on real-world edge devices such as a CPU when the pre-trained model is large (such as an LLM). Formally, given a test
sample $X_{test}$, our goal is to find an optimal prompt $\boldsymbol{p}^*$:

\begin{equation}  
    \boldsymbol{p}^*=\arg\min_{\boldsymbol{p}} 
     \mathcal{L}_{\mathrm{PGTPT}}(\mathcal{F},\boldsymbol{p},X_\mathrm{test})
     \label{ppa}
\end{equation}

using the physics-guided self-supervised loss $\mathcal{L}_{\mathrm{PGTPT}}$ from Equation \ref{ecm}. Here, $\mathcal{F}$ denotes the pre-trained model. 
We will also demonstrate experimentally that PPA significantly reduces the cost of TTA while only marginally reducing its performance.

\subsection{Integrete Existing SOH Estimation Models into BatteryTTT} \label{3.5}

In this subsection, we demonstrate how to incorporate existing SOH estimation algorithms \cite{zhangbatteryml} into our test-time adaptation framework to complete the whole cross-domain SOH estimation process depicted in Figure \ref{overview}. 
Existing SOH estimation studies \cite{roman2021machine, zhangbatteryml} extensively use statistical models and high-performance neural networks for LIB SOH estimation, such as Gaussian Process Regression \cite{williams2006gaussian}, Random Forest \cite{geurts2006extremely}, MLP \cite{haykin1998neural}, RNNs\cite{hochreiter1997long} and Transformer \cite{vaswani2017attention} for SOH estimation. Only deep learning methods can be incorporated into the TTA framework, as well as other transfer learning settings.

Specifically, our architecture follows a Y-shaped design, as described in \cite{shu2022test, gandelsman2022test}: a feature extractor $f$ is simultaneously followed by a self-supervised head $g$ and a main task head $h$. Here, we substitute $f$ with the encoder of existing neural networks for SOH estimation, such as GRU, LSTM, and Transformer, and $g$ with the decoder. For the main regression task head $h$ for SOH estimation, we use a linear projection from the dimension of the encoder features to 1, which is primarily a historic artifact \cite{he2022masked}.

During pre-training, we first train $g\circ f$ using the PG-SSL loss in Equation \ref{pg-ssl}  in an unsupervised manner with the features of the source LIB dataset. Then we perform linear probing by combining the encoder $f$ with the main task head $h$, keeping $f$ frozen. Formally:

\begin{equation}f_\mathbf{0},\boldsymbol{g_0}=\arg \min_{\boldsymbol{f},\boldsymbol{g}}\frac1n\sum_{i=1}^n\mathcal{L}_{\mathrm{PG-SSL}}(x_i;\boldsymbol{f},\boldsymbol{g})\end{equation}

Followed by:

\begin{equation}h_0=\arg\min_h\frac1n\sum_{i=1}^n\mathcal{L}_{\mathrm{MSE}}(h\circ f_0(x_i),y_i)\end{equation}

The summation is over the training set with $n$ samples, each consisting of input $x_{i}$ and label $y_{i}$.
During test-time adaptation, we optimize $g\circ f$ before making a prediction each time a test input $x$ arrives. After optimization, we make a prediction on $x$ as $h \circ f_x(x)$, formally as:

\begin{equation}f_x,g_x=\arg\min_{f,g}\mathcal{L}_{\mathrm{PG-SSL}}(x;f_0,g_0)
\label{tta}
\end{equation}

\begin{equation}prediction =  h_0\circ f_x(x)\end{equation}

In particular, we argue for the use of our proposed prefix prompt adaptation in Equation \ref{ppa} to avoid the substantial computation brought by fine-tuning Equation \ref{tta} at inference time.






\subsection{Integrate an LLM into BatteryTTT (GPT4Battery)}

 In this subsection, we explore the potential of LLM in battery sequence modeling. Inspired by recent studies which utilize an LLM for protein sequence prediction \cite{lv2024prollama} and time series analysis \cite{jin2023time}, we primarily employ the idea of \emph{model reprogramming} to effectively align the modalities of battery data and natural language, leveraging the reasoning and generalization abilities of LLMs for battery tasks.

Specifically, we first tokenize and map the input Qdlinear features into a high-dimensional space with the same dimensionality as the word space of the language model. Then, we fuse the information of battery sequence modality and language modality using a cross-attention layer.

However, for a word embedding space of an LLM $\mathbf{E}\in\mathbb{R}^{V\times D}$, where $V$ is the vocabulary size. The vocabulary size can be inevitably large (for example, GPT2 has a V of 50257 \cite{radford2019language}). Simply leveraging $\mathbf{E}$ will result in large and potentially dense reprogramming space, increasing the computation complexity and difficulty of catching the relevant source tokens. Following \cite{sun2023test} and \cite{jin2023time}, we maintain only a small collection of text prototypes by linearly probing $\mathbf{E}$, denoted as $\mathbf{E^{\prime}}\in\mathbb{R}^{V^{\prime}\times D}$, where $V^{\prime} \ll V$. 
Then, we align the tokenized input patches and text prototypes with a multi-head cross-attention layer. Specifically:

\begin{equation*}
\begin{aligned}
   \mathbf{Z}_k^{(i)} &=\text{attention} (\mathbf{Q}_k^{(i)},\mathbf{K}_k^{(i)},\mathbf{V}_k^{(i)})\\
   &=\text{softmax}(\frac{\mathbf{Q}_k^{(i)}\mathbf{K}_k^{(i)\top}}{\sqrt{d_k}})\mathbf{V}_k^{(i)} 
\end{aligned}
\end{equation*}

By aggregating each $\mathbf{Z}_{k}^{(i)}\in\mathbb{R}^{P\times d}$ in every head, we obtain  $\mathbf{Z}^{(i)}\in\mathbb{R}^{P\times D}$.
This way, the text prototypes can learn cues in language which can then represent the relevant local patch information. We will experimentally demonstrate the improvement in generalizability achieved by incorporating LLMs and the effectiveness of model reprogramming.



\section{Experiments}\label{sec:experiments}

\begin{table*}
    \centering
    \caption{Main specifications of selected LIB datasets.}
    
    \resizebox{0.85\linewidth}{!}{%
    \begin{tabular}{lp{1.8cm}p{1.7cm}llp{2.5cm}l}
        \hline
        \textbf{Dataset}   & \makecell{\textbf{Electrode  Material}} & \makecell{\textbf{Nominal Capacity} } & \makecell{\textbf{Voltage Range }} & \textbf{Samples} & 
        \textbf{Estimated Collect Time} &
        \textbf{Collector}    \\
        \hline                                     
        CALCE       &LCO & 1.1 (Ah) & 2.7-4.2 (V) & 2807 &1397 (hour)& University of Maryland \\
        
        SANYO       & NMC & 1.85 (Ah) & 3.0-4.1 (V) & 415 & 644 (hour) & RWTH Aachen University \\
        
        KOKAM        &LCO/NCO & 0.74 (Ah) & 2.7-4.2 (V) &503 & 8473 (hour) &University of Oxford \\
        
        PANASONIC     &NCA &3.03 (Ah) &2.5-4.29 (V) &2770 &1801 (hour)& Beijing Institude of Technology \\
        
        GOTION         &LFP &27 (Ah) &2.0-3.65 (V) & 4262 &2238 (hour)& Beijing Institude of Technology \\

        NHRY & LFP & \textbf{300 (Ah)} & 2.5-3.5 (V) & 808 & 1200 (hour) & \textbf{Ours} \\
        \hline
    \end{tabular}
    }
    \label{datasets}
\end{table*}

In this section, we empirically evaluate the proposed approach on six real-world LIB datasets, including five publicly available datasets for daily applications (with capacities ranging from a few Ah) and our own collected 300Ah large LIB dataset for energy storage. Specifically, we focus on (1) the overall improved generalization performance of TTA and the superior performance by combining GPT-2 and TTA (GPT4Battery), (2) an efficacy study of the two proposed designs, PG-SSL and PPA, (3) the ablation results of GPT4Battery, (4) inference efficiency of involving TTA, and (5) the scalability and deployment of our method on large energy storage LIBs.

\subsection{Experiment Settings}
\subsubsection{Dataset preparation.}
We conducted experiments using five publicly available lithium-ion battery (LIB) datasets intended for daily commercial use, with capacities ranging from 0.74 Ah to 27 Ah. Additionally, we utilized four of our own LIB datasets collected for industrial-level energy storage, specifically with a capacity of 300 Ah, which we will also make publicly available for academic purposes. These datasets encompass a variety of widely used cathode active materials, capacities, and manufacturers. A summary of the dataset statistics is presented in Table \ref{datasets}.

\subsubsection{Baselines} 
We compare our method with four types of baselines to demonstrate the efficacy of the proposed BatteryTTT framework and the GPT4Battery model: (1) Existing non-transfer learning machine learning (ML) methods for LIB State of Health (SOH) estimation, including Gaussian Process Regression \cite{williams2006gaussian}, Random Forest \cite{geurts2006extremely}, Multi-Layer Perceptron (MLP) \cite{haykin1998neural}, Recurrent Neural Networks (RNNs) \cite{hochreiter1997long}, and Transformer \cite{vaswani2017attention}\footnote{BatteryML \cite{zhangbatteryml} provides a comprehensive platform summarizing these models}; (2) Integration of existing models into our BatteryTTT framework; (3) State-of-the-art transfer learning methods for cross-domain LIB SOH estimation, such as woDegrad \cite{lu2023deep} and SSF-WL \cite{wang2024lithium}\footnote{These methods rely on sufficient unlabeled target data (UTDs) to operate effectively, which requires impractical data collection time, as summarized in Table \ref{assumptions}.}; and (4) Integration of large language models (LLMs) into the BatteryTTT framework (GPT4Battery).

We reproduce the results of BatteryML and woDegrad based on the provided code and follow the approach details for SSF-WL. For a fair comparison, we adhere to the data pre-processing methods outlined in \cite{zhangbatteryml} and use \emph{QdLinear} \cite{attia2021statistical} as the unified feature set. We adopt the standard evaluation metrics of mean absolute error (MAE) and root mean squared error (RMSE).

\subsubsection{Implementation details}
Our models are implemented using Pytorch and trained on a single 3070Ti GPU. 
We utilize the AdamW optimizer \cite{loshchilov2017decoupled} with a fixed learning rate of 1e-3 for pre-training and linear probing until convergence. The mask ratio is set to 30\% during this phase.
TTA is conducted using stochastic gradient descent (SGD) with a momentum of 0.9 due to its consistency in improving performance on distribution shifts \cite{gandelsman2022test}. Typically, we set a fixed learning rate of 1e-2 and iterate for 10 steps, as more steps only marginally improve performance based on our observations. The mask ratio during TTA will be specifically analyzed later.
For further details, we have made the relevant code and data available at the following link: \url{https://anonymous.4open.science/r/gpt4battery-55FC}.


\subsection{Main Performance}
In this section, we report the main improvements in cross-domain generalization performance of our proposed TTA methods (e.g. PG-SSL and PPA), on five commercial LIB datasets for daily usage. Additionally, we demonstrate that by leveraging an LLM as the backbone, GPT4Battery achieves superior generalization performance compared to all baseline methods.

\subsubsection{Improvement of Generalization Ability of TTA}

\begin{table}[htbp]
  \centering
  
  \caption{Improved performance of TTA on existing methods and comparison with current transfer learning methods.}
  \resizebox{\columnwidth}{!}{
    \begin{tabular}{c|cc|cc|cc|cc|cc}
    \toprule
    \multirow{2}[4]{*}{Models} & \multicolumn{2}{c|}{CALCE} & \multicolumn{2}{c|}{SANYO} & \multicolumn{2}{c|}{KOKAM} & \multicolumn{2}{c|}{PANASONIC} & \multicolumn{2}{c}{GOTION} \\
\cmidrule{2-11}          & MAE   & RMSE  & MAE   & RMSE  & MAE   & RMSE  & MAE   & RMSE  & MAE   & RMSE \\
    \midrule
    Random Forest & 5.76  & 4.9   & 7.87  & 6.73  & 5.76  & 4.88  & 4.96  & 4     & 0.62  & 0.54 \\
    Light GBM & 6.8   & 5.42  & 6.31  & 5.62  & 6.52  & 5.31  & 6.06  & 4.97  & 0.55  & 0.47 \\
    MLP   & 3.93  & 3.52  & 6.1   & 5.93  & 13.1  & 10.9  & 5.08  & 4.38  & 0.54  & \textbf{0.44} \\
    GRU   & 2.18  & 2.86  & 9.17  & 9.19  & 3.07  & 3.44  & 2.53  & 3.5   & 0.74  & 0.84 \\
    LSTM  & 2.52  & 2.78  & 7.58  & 7.84  & 3.07  & 3.93  & 1.55  & 2.19  & 1.65  & 1.68 \\
    Transformer & 2.27  & 2.57  & 8.1   & 8.24  & 15.3  & 17.1  & 1.9   & 2.35  & 1.28  & 1.4 \\
    \midrule
    GRU+ (TTA)   & 1.9   & 2.25  & 7.8   & 8.01  & \textbf{2.4} & \textbf{2.74} & 1.74  & 2.73  & 0.67  & 0.71 \\
    LSTM+ & 2.08  & 2.36  & 6.71  & 7.04  & 2.93  & 3.67  & 1.31  & 2     & 0.65  & 0.77 \\
    Transformer + & 1.83  & 1.97  & 7.03  & 7.18  & 13.2  & 14.6  & 1.32  & \textcolor[rgb]{ 1,  0,  0}{\textbf{1.85}} & \textcolor[rgb]{ 1,  0,  0}{\textbf{0.34}} & \textcolor[rgb]{ 1,  0,  0}{\textbf{0.43}} \\
    GPT-2+  & \textcolor[rgb]{ 1,  0,  0}{\textbf{1.52}} & \textcolor[rgb]{ 1,  0,  0}{\textbf{1.89}} & 1.35  & 1.71  & 7.95  & 8.01  & \textcolor[rgb]{ 1,  0,  0}{\textbf{1.28}} & \textbf{1.95} & \textbf{0.38} & 0.47 \\
    Bert+  & 2.01  & 2.33  & 1.46  & 1.82  & 7.77  & 8.04  & 1.52  & 2.2   & 0.41  & 0.52 \\
    Llama-7b+ & 1.57  & 1.94  & 1.35  & 1.69  & 8.58  & 9 .66 & \textbf{1.3} & 2.01  & 0.4   & 0.49 \\
    \midrule
    woDegrad & 1.76  & 1.96  & \textbf{1.21} & \textbf{1.54} & \textcolor[rgb]{ 1,  0,  0}{\textbf{1.76}} & \textcolor[rgb]{ 1,  0,  0}{\textbf{3.01}} & 2.09  & 2.56  & 0.45  & 0.58 \\
    SSF-WL & \textbf{1.55} & \textbf{1.93} & \textcolor[rgb]{ 1,  0,  0}{\textbf{1.08}} & \textcolor[rgb]{ 1,  0,  0}{\textbf{1.24}} & 6.21  & 5.1   & 1.44  & 2.06  & 0.51  & 0.72 \\
    \bottomrule
    \end{tabular}%
    }
  \label{main}%
\end{table}%

Table \ref{main} shows the main improvements in the generalization performance of our proposed TTA methods. We use the GOTION dataset as the source dataset for its extensive label coverage and then include its own test set along with the remaining four datasets (CALCE, SANYO, KOKAM and PANASONIC) as the target datasets for generalizability testing. Overall, we observe that the TTA method shows a significant performance improvement of about 50\% compared to the no-TTA method. Some models (such as Transformer, GPT2) equipped with TTA achieve a performance that rivals or even exceeds the performance of current transfer learning methods that require additional access to the target data in CALCE, PANASONIC and GOTION. Specifically, within all the methods that use TTA, the utilization of large language models also has made quite a difference in generalization performance improvement. For instance, GPT2+ achieves first or second rank performance on the CALCE, GOTION and PANASONIC datasets compared to all baseline methods. 
LLama+ also obtained the best and second-best performance on the SANYO and PANASONIC dataset, respectively, compared to the other TTA methods.
We also observe that some model architectures dominate the performance on certain datasets over TTA, e.g., the RNN family (GRU, LSTM) outperforms the Transformer family on KOKAM as a whole, and the TTA-enhanced GRU+ achieves performance comparable to that of woDegrad.


\subsubsection{Superior Performance of GPT4Battery over all Baseline Methods}

\begin{figure}[h]
  \centering
    \includegraphics[width=0.85\linewidth]{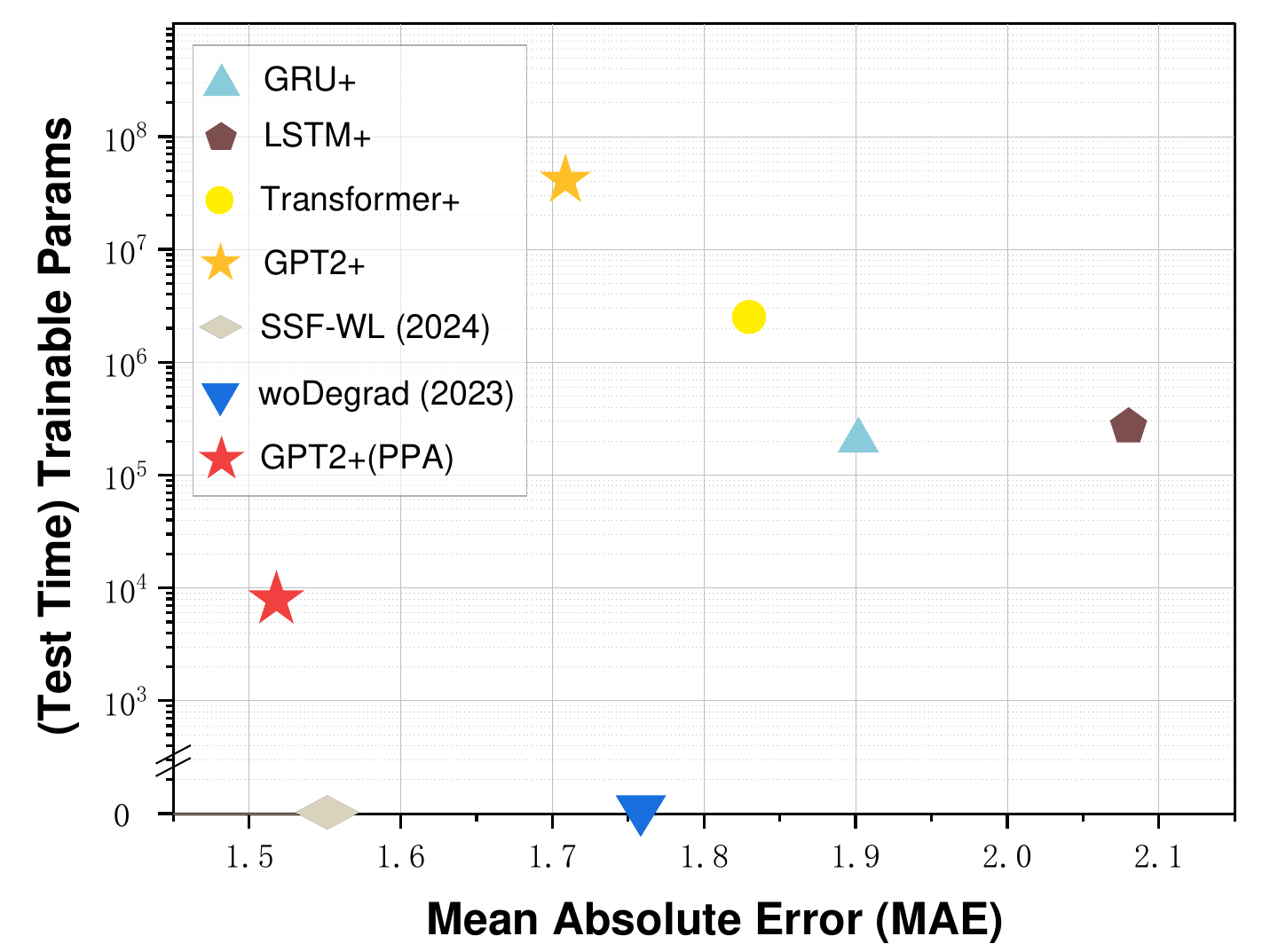}
  \caption{Superior Performance of GPT4Battery (on CALCE dataset).}
  \label{gpt4battery}
\end{figure}


In this section, we provide a detailed comparison of the generalizability gains and computation trade-offs achieved by incorporating LLMs. Figure \ref{gpt4battery} demonstrates that GPT2 equipped with TTA (which we name GPT4Battery) obtains the lowest MAE results on the CALCE dataset than all other models including current transfer learning methods (woDegrad and SSF-WL) that require additional assumptions to access the target dataset. GPT4battery also gets very competitive results in other datasets as do other large models (e.g. Llama and Bert).

The results of applying Prefix Prompt Adaptation (PPA) to large language models (LLMs), as illustrated in Figure \ref{gpt4battery}, are particularly noteworthy. Conventional fine-tuning of GPT-2's positional encoding and layer normalization significantly increased the number of trained parameters at test time, by a factor of 10 to 100 compared to Transformer+ and RNNs+, while only (relatively) slightly reducing the mean absolute error (MAE) from 1.83 to 1.71. In contrast, our Prefix Prompt Adaptation (PPA) approach not only reduces the number of adjustable parameters by nearly 10,000 times, making GPT4Battery ten times more parameter-efficient than the RNN series, but also further reduces the MAE from 1.71 to 1.52, an improvement of approximately 10.7\%. However, we should note the limitation that involving an LLM does increase the inference time even with PPA, which is a trade-off between accuracy and computation efficiency.

It is important to note that while applying PPA to regular Transformer+ and RNNs+ also reduces the number of trainable parameters to the order of $10^3$, it can impair their generalization performance, resulting in a slightly inferior performance compared to full-parameter fine-tuning. Therefore, we overall report the best performance for their full parameter fine-tuning, while reporting the best performance for LLMs using PPA in Table \ref{main} and Figure \ref{gpt4battery}. The performance of other models using PPA is thoroughly ablated in Sec. \ref{4.3.1}.


\subsection{Efficacy Analysis}
In this section, we analyze the effectiveness and important design choices of each component. Specifically, we evaluate the two TTA designs: Physical-Guided Self-supervised Learning (PG-SSL) and Prefix Prompt Adaptation (PPA).

\subsubsection{Effect and Computation Trade-offs of Prefix Prompt Adaptation} \label{4.3.1}

\begin{figure}[htbp]
    \centering
    \includegraphics[width=\columnwidth]{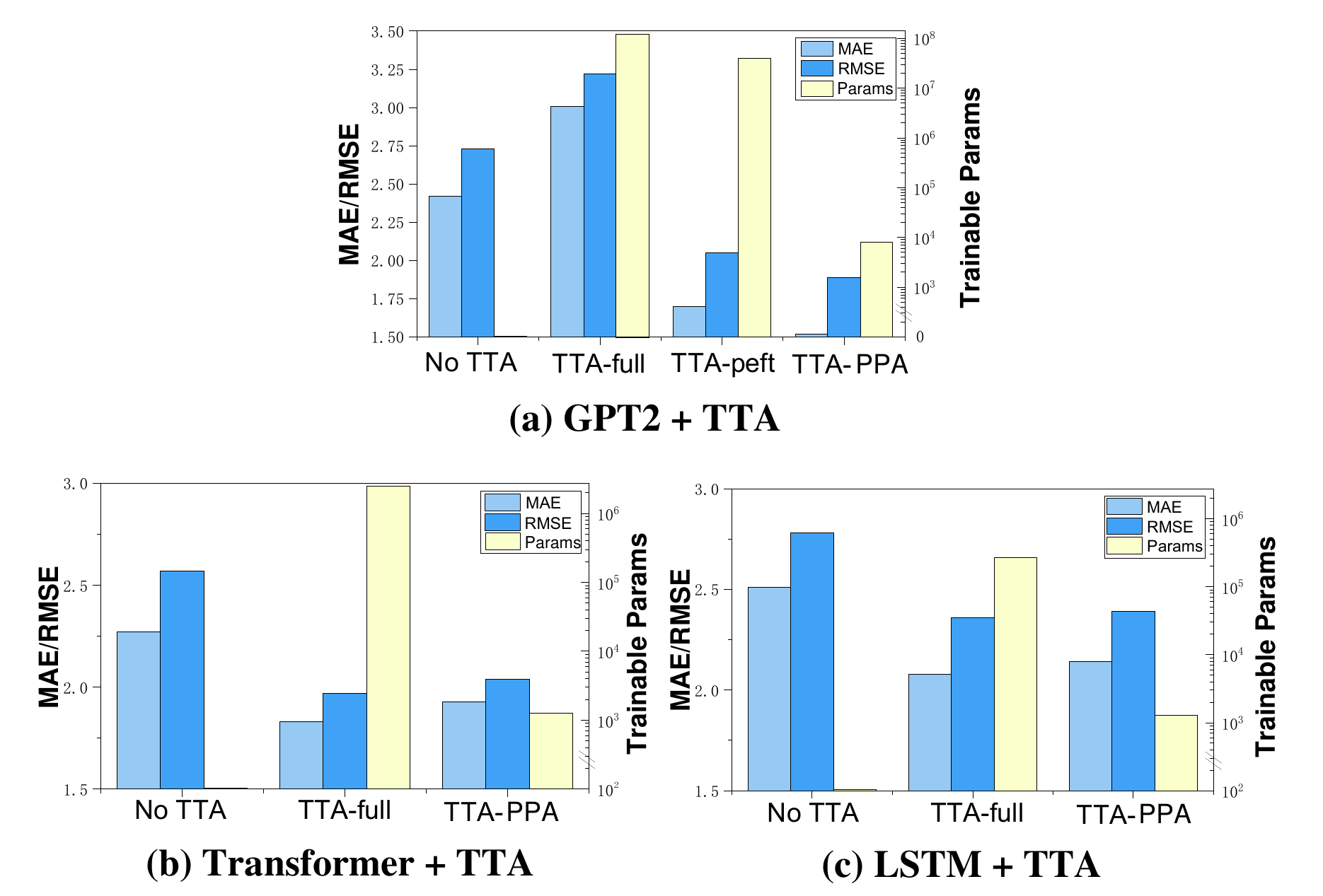} 
    \caption{Efficacy and computation trade-offs of Prefix Prompt Adaptation (PPA) on different models.} 
    \label{PPA} 
\end{figure}

By adjusting only a small number of soft
prompts prefix to the input context, this design can significantly reduce the adjustable parameters to $10^3$ orders of magnitude regardless of model size, as illustrated in Figure \ref{PPA} on the GPT2, Transformer and LSTM models. Applying PPA to Transformer and LSTM slightly impact the accuracy, as shown in Figure \ref{PPA}. Applying PPA to LLMs however, significantly reduces the Mean Absolute Error (MAE) from 3.89 to 1.52, a reduction of approximately 60.9\%. Moreover, PPA achieves better generalization performance than traditional parameter-efficient fine-tuning of the positional encoding and layer normalization layers \cite{zhou2023one} of the LLM. We attribute this to the language-agnostic pattern recognition and inference capabilities acquired through pre-training on text corpora \cite{gruver2024large, mirchandani2023large}. Guided by learnable prefix prompts, these capabilities can also be generalized to battery sequence data.

\subsubsection{Efficacy and Parameter Sensitivity of Physical-Guided Self-supervised Learning}

In this section, we analyze the effectiveness of PG-SSL and conduct a sensitivity analysis of another important parameter affecting the self-supervised loss, the mask ratio. We performed ablation studies on losses using physical constraints and pure mask reconstruction. Additionally, we analyzed the effect of different masking rates on the MAE reduction of TTAs. We employ the transformer model and full-parameter tuning for TTA, conducting experiments on the adaptation to the CALCE and PANASONIC datasets.

\begin{figure}[htbp]
    \centering
    \includegraphics[width=1\columnwidth]{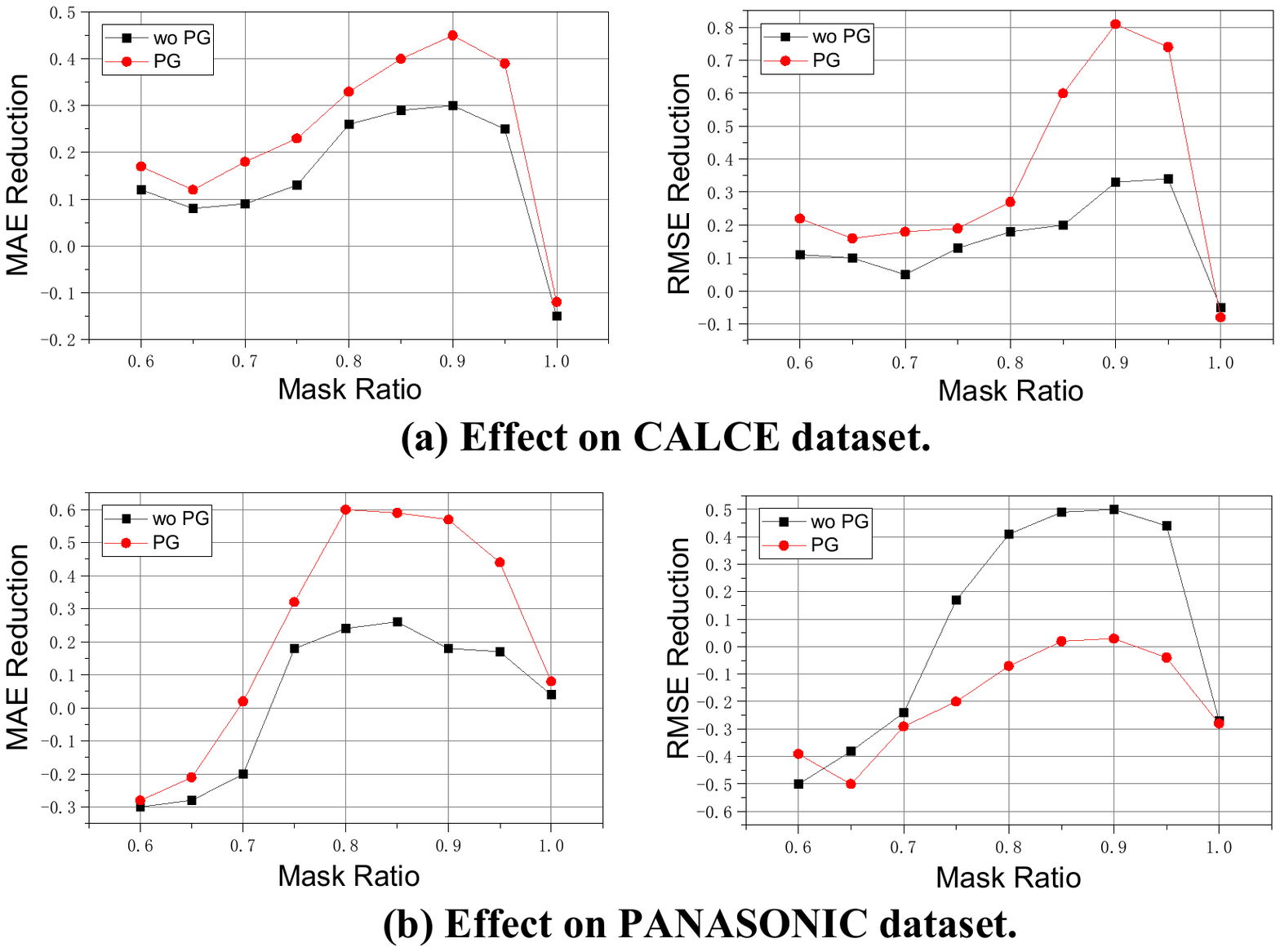} 
    \caption{Effect on MAE/RMSE Reduction of Physical-Guided Self-supervised Learning and Sensitivity of Mask Ratio.} 
    \label{PG} 
\end{figure}

Figure \ref{PG} shows that the inclusion of the physical-guided loss seamlessly enhances the performance in both MAE and RMSE reduction. Comparatively, the MAE reduction of using PG-SSL is slightly more significant in the CALCE data set (Fig. 6 (a)) than in the PANASONIC data set (Figure \ref{PG} (b)). This is because the former is a much more nonlinear dataset, making pure reconstruction loss less effective in capturing representations.
We also observe from both (a) and (b) of Figure \ref{PG} that a larger mask ratio of 0.7 to 0.9 promotes learning a better representation with or without PG-SSL, while a small mask ratio
of 0.5 to 0.6 fails. This is consistent with the MAE-based self-supervised learning observations in the computer vision domain, where masking a high proportion of the input image, e.g., 75\%, yields a non-trivial and meaningful self-supervisory task. In summary, the combination of masking input and PG-SSL results in the best TTA results.

\subsection{Ablation Study of GPT4Battery}

In this section, we consider the best performing GPT2+TTA as a self-contained model (GPT4Battery) and provide ablation study on the effects of different components or design choices. 
Our results in Table \ref{ablation} indicate that ablating either model reprogramming or any other designs in test-time adaptation hurts the generalization performance on unseen LIBs. In the absence of physical guidance, we observe a notable average performance degradation of \textbf{55.1\%}, which becomes more pronounced (i.e., exceeding $\textbf{70\%}$) when discarding the TTA strategy completely. The act of model reprogramming also stands as a pivotal element in cross-modality alignment, enabling the LLM to understand the LIB's sequence data with the help of text prototypes. Ablation of reprogramming results leads to over \textbf{20\%} degradation on average performance. 

\begin{table}[htbp]
  \centering
  \small
  \caption{Ablating the different components of GPT4Battery. \textcolor[rgb]{1,  0,0}{\textbf{Red:}} the best.}
  \resizebox{\columnwidth}{!}{
    \begin{tabular}{lccccc}
    \midrule
    Method & CALCE & SANYO & KOKAM & PANAS. & GOTION \\
    \midrule
    GPT4Battery & \textcolor[rgb]{1,0,0}{\textbf{1.52}}  & \textcolor[rgb]{1,0,0}{\textbf{1.35}}  & \textcolor[rgb]{1,0,0}{\textbf{7.95}}     & \textcolor[rgb]{1,0,0}{\textbf{1.28}}  & \textcolor[rgb]{1,0,0}{\textbf{0.25}}  \\
    w/o PG-SSL & 2.11  & 1.12  & 8.34  & 3.11  & 0.265 \\
    w/o Masked TTA & 2.05  & 0.97  & 8.44  & 1.22  & 0.255  \\
    w/o TTA & 2.13  & 1.34  & 9.51  & 3.44  & 0.297 \\
    w/o model reprogramming  & 4.13  & 2.78  & 10.56  & 4.57  & 0.31 \\
    
    \midrule
    \end{tabular}%
    }
  \label{ablation}%
\end{table}%

\subsection{TTA Efficiency}
We present the average running time across five datasets for three representative models equipped with TTA and two current transfer learning methods. Our focus is primarily on model inference time, as TTA does not significantly impact training duration. Table \ref{effiency} demonstrates that although TTA-equipped methods achieve substantially higher accuracy, they are 10 to 100 times slower than existing methods. This trade-off highlights that our approach sacrifices some speed for enhanced accuracy. However, the resulting $10^2$ speed level remains sufficient for modern BMS requirements, where individual response times typically need to be under 500ms in our deployed system. More importantly, our framework can eliminate the need for 644 to 8,473 hours of degradation experiments for data collection, as summarized in Tables \ref{assumptions} and \ref{datasets}.

\begin{table}[htbp]
  \centering
  \caption{Efficiency Analysis of TTA.}
  \resizebox{\columnwidth}{!}{
    \begin{tabular}{cccccr}
    \toprule
    Method & GPT2+ & Transformer+ & LSTM+ & woDegrad & \multicolumn{1}{c}{SSF-WL} \\
    \midrule
    Overall Inference Time (s) & 32.6  & 6.1   & 4.3   & 0.35  & \multicolumn{1}{c}{0.3} \\
    \multicolumn{1}{l}{One Inference Time (ms)} & 51.17 & 9.57  & 6.75  & 0.55  & \multicolumn{1}{c}{0.47} \\
    Model parameters & 7680  & 1280  & 1280  & 68774000 & 2601369 \\
    \bottomrule
    \end{tabular}%
    }
  \label{effiency}%
\end{table}%


\subsection{Scalability on Large Energy Storage LIBs}

In this section, we report the scalability of the proposed framework through deployment on our battery management system (BMS) for industry-level energy storage LIBs (300Ah).

\subsubsection{LIB Dataset}

We collect the NHRY dataset by performing degradation experiments on 8 industry-level energy storage LIBs (300Ah) from 4 brands (Ningde, Haichen, Ruipu, Yiwei). Test environment and procedures comply with the China Standard BMS for Energy Storage GB/T 34131-2023. A total of 800 degradation cycles were experienced ranging over two months from December 2022 to January 2023.

\subsubsection{Online Performance}

For four different brands of LIBs with the same capacity, we mix two of the brands as the source domain and evaluate the online performance of the LIBs from the remaining two brands, resulting in a total of two combinations for the cross-battery setting. We report the MAE/RMSE metric along with the inference time (ms) using representative baselines.

\begin{table}[htbp]
  \centering
  \caption{Online performance on Large Energy Storage LIBs.}
  \resizebox{\columnwidth}{!}{
    \begin{tabular}{cccc|ccc}
    \toprule
    \multicolumn{1}{c}{} & \multicolumn{3}{c|}{Combination 1} & \multicolumn{3}{c}{Combination 2} \\
    \midrule
    \multicolumn{1}{c}{} & \multicolumn{1}{c}{MAE} & \multicolumn{1}{c}{RMSE} & \multicolumn{1}{l|}{Per Infer. Time} & \multicolumn{1}{c}{MAE} & \multicolumn{1}{c}{RMSE} & \multicolumn{1}{l}{Per Infer. Time} \\
    \midrule
    LSTM  & 4.33  & 4.59  & 4.23  & 2.44  & 2.69  & 3.23 \\
    Transformer & 5.62  & 5.89  & 7.61  & 3.48  & 3.78  & 6.32 \\
    LSTM+ & 0.775 & 0.788 & 54.53 & 0.55  & 0.62  & 44.85 \\
    Transformer+ & 1.25  & 1.38  & 88.57 & 1.02  & 1.23  & 77.83 \\
    GPT4Battery & 0.734 & 0.761 & 103.47 & 0.314 & 0.418 & 105.68 \\
    \bottomrule
    \end{tabular}%
    }
  \label{tab:addlabel}%
\end{table}%

\section{Conclusion}\label{sec:conclusion}

In this paper, we propose a novel test-time adaptation (TTA) framework for cross-domain LIB state of health (SOH) estimation. This one sample adaptation setting allows the model to continually adapt to the target domain with every single unlabeled test sample, perfectly aligning with the nature of battery degradation features, which can only be obtained one by one during the long aging process. This setting also addresses the limitations of existing transfer learning methods, which assume additional access to the target LIB dataset, thereby saving months to years of labor in data collection.
By introducing GPT4Battery, we hope this work will inspire the research community to fully leverage advanced AI techniques, such as Test-Time Adaptation (TTA) and large language models (LLMs), for more AI4Science problems, thereby saving enormous time and accelerating scientific discovery.


\newpage

\balance
\bibliographystyle{ACM-Reference-Format}
\bibliography{reference}

\newpage
\appendix

\begin{table*}[htbp]
  \centering
  \caption{Degradation Conditions of the Selected LIB Datasets}
    \begin{tabular}{cp{6em}p{40em}}
    \toprule
    \multicolumn{1}{c}{Dataset} & \multicolumn{1}{c}{Cell Information} & \multicolumn{1}{c}{Test Conditions} \\
    \midrule
    CALCE & 3 cells termed CS2-35, CS2- 36, CS2-37 & $\bullet$ Charged at a constant current rate of 0.5C until the voltage reached 4.2V and then 4.2V was sustained until the charging current dropped to below 0.05A. \newline{} $\bullet$ Discharged at a constant current rate of 1C. \newline{} $\bullet$ Ambient temperature is not mentioned. \\

    \midrule
    SANYO & 48 commercial
UR18650E
cylindrical
cells & $\bullet$ Charged at a constant current rate of
1C until the voltage reached 4.1V and
then 4.1V was sustained until the
charging current dropped to below
0.04A. \newline{} $\bullet$ Discharged at a constant current rate
of 1C. \newline{} $\bullet$ 25°C.\\

    \midrule
    KOKAM & 8 KOKAM
SLPB533459H
4 pouch cells & $\bullet$ Charged at a constant current rate of
1C until the voltage reached 4.2V. \newline{} $\bullet$ Discharged at a constant current rate of 1C. \newline{} $\bullet$ 40°C. \\

    \midrule
    PANASONIC & 3 commercial
NCR18650BD
cylindrical
cells & $\bullet$ Charged at a constant current rate of
0.3C until the voltage reached 4.2V
and then 4.2V was sustained until the
charging current dropped to below
0.03A. \newline{} $\bullet$ Discharged at a constant current rate of 2C. \newline{} $\bullet$ 20°C.
 \\

    \midrule
    GOTION & 3 commercial
IFP20100140A
cells
&Charged at a constant current rate of
1C until the voltage reached 3.65V
and then 3.65V was sustained until the
charging current dropped to below
1.35A.
 \newline{} $\bullet$ Discharged at a constant current rate of 1C. \newline{} $\bullet$ 45°C. \\
 \bottomrule
 \label{condition}%
    \end{tabular}%
\end{table*}%

\section{Appendix: Pseudo Code}

\begin{algorithm}
    \caption{Pretraining and Test-Time Adaptation of BatteryTTT}
    \begin{algorithmic}[1]
        \STATE \textbf{Input:} Source LIB dataset $(x_i, y_i) \text{ for } i = 1, \ldots, n$
        \STATE \textbf{Output:} Prediction for test input $x$
        
        \STATE \textbf{Pre-training Phase:}
        \FOR{$i = 1$ \TO $n$}
            \STATE Compute $\mathcal{L}_{\mathrm{PG-SSL}}(x_i; \boldsymbol{f}, \boldsymbol{g})$ using Equation \ref{pg-ssl}
        \ENDFOR
        \STATE \textbf{Train:} $f_{\mathbf{0}}, g_0 = \arg \min_{\boldsymbol{f}, \boldsymbol{g}} \frac{1}{n} \sum_{i=1}^n \mathcal{L}_{\mathrm{PG-SSL}}(x_i; \boldsymbol{f}, \boldsymbol{g})$ (Equation 1)

        \STATE \textbf{Linear Probing:}
        \FOR{$i = 1$ \TO $n$}
            \STATE Compute $\mathcal{L}_{\mathrm{MSE}}(h \circ f_0(x_i), y_i)$
        \ENDFOR
        \STATE \textbf{Train:} $h_0 = \arg \min_h \frac{1}{n} \sum_{i=1}^n \mathcal{L}_{\mathrm{MSE}}(h \circ f_0(x_i), y_i)$ (Equation 2)

        \STATE \textbf{Test-Time Adaptation Phase:}
        \FOR{each test input $x$}
            \STATE Optimize $f_x, g_x = \arg \min_{f,g} \mathcal{L}_{\mathrm{PG-SSL}}(x; f_0, g_0)$ (Equation \ref{tta})
            \STATE Compute $prediction = h_0 \circ f_x(x)$ (Equation 4)
        \ENDFOR

        \STATE \textbf{Prefix Prompt Adaptation:}
        \STATE Use proposed prefix prompt adaptation to avoid computation during inference
    \end{algorithmic}
\end{algorithm}

\section{Appendix: Degradation Conditions of Selected Datasets}
Table \ref{condition} provides the specific degradation conditions of the selected lithium-ion batteries (LIBs) used in this work, including detailed cell information and the protocols for charging and discharging.




\end{document}